\documentclass[runningheads]{llncs}
\usepackage[T1]{fontenc}
\usepackage{graphicx,verbatim}
\usepackage{multirow} 
\usepackage[colorlinks=true,allcolors=blue]{hyperref} 
\usepackage[misc]{ifsym}
\usepackage[normalem]{ulem} 
\useunder{\uline}{\ul}{}
\usepackage{amsmath,amssymb,amsfonts,bm} 

\usepackage[table,xcdraw]{xcolor}
\usepackage{floatrow} 
\usepackage{booktabs} 
\usepackage{placeins}
\newfloatcommand{capbtabbox}{table}[][\FBwidth]
\begin{document}
%
%\titlerunning{TAT: Task-Adaptive Transformer for All-in-One Medical Image Restoration}
\title{TAT: Task-Adaptive Transformer for All-in-One Medical Image Restoration}

\author{
Zhiwen Yang\inst{1} \and
Jiaju Zhang\inst{1} \and 
Yang Yi\inst{1} \and 
Jian Liang\inst{2} \and 
Bingzheng Wei\inst{3} \and 
Yan Xu\inst{1}\textsuperscript{(\Letter)}
} 

% 1{Yang, Zhiwen}, 2{Zhang, Jiaju}, 3{Yi, Yang}, 4{Liang, Jian}, 5{Wei, Bingzheng}, 6{Xu, Yan}

\authorrunning{Yang et al.}
\titlerunning{TAT: Task-Adaptive Transformer}
\institute{
School of Biological Science and Medical Engineering, State Key Laboratory of Software Development Environment, Key Laboratory of Biomechanics and Mechanobiology of Ministry of Education, Beijing Advanced Innovation Center for Biomedical Engineering, Beihang University, Beijing 100191, China
\\
\email{xuyan04@gmail.com} \and 
Sinovision Technologies (Beijing) Co., Ltd., Beijing 101102, China \and
ByteDance Inc., Beijing 100098, China
}

\maketitle      
\begingroup
  \renewcommand\thefootnote{}\footnote{%
 Equal contribution -- Z. Yang, J. Zhang, and Y. Yi.
  } 
\endgroup

\begin{abstract}
Medical image restoration (MedIR) aims to recover high-quality medical images from their low-quality counterparts. Recent advancements in MedIR have focused on All-in-One models capable of simultaneously addressing multiple different MedIR tasks. However, due to significant differences in both modality and degradation types, using a shared model for these diverse tasks requires careful consideration of two critical inter-task relationships: \textit{task interference}, which occurs when conflicting gradient update directions arise across tasks on the same parameter, and \textit{task imbalance}, which refers to uneven optimization caused by varying learning difficulties inherent to each task. To address these challenges, we propose a task-adaptive Transformer (TAT), a novel framework that dynamically adapts to different tasks through two key innovations. First, a task-adaptive weight generation strategy is introduced to mitigate \textit{task interference} by generating task-specific weight parameters for each task, thereby eliminating potential gradient conflicts on shared weight parameters. Second, a task-adaptive loss balancing strategy is introduced to dynamically adjust loss weights based on task-specific learning difficulties, preventing task domination or undertraining. Extensive experiments demonstrate that our proposed TAT achieves state-of-the-art performance in three MedIR tasks—PET synthesis, CT denoising, and MRI super-resolution—both in task-specific and All-in-One settings. Code is available at \href{https://github.com/Yaziwel/TAT}{this https URL}.

% Authors must provide keywords and are not allowed to remove this Keyword section.

\end{abstract}
\section{Introduction} 
% Medical image restoration (MedIR) is a fundamental task in medical imaging, focused on reconstructing high-quality (HQ) images from their low-quality (LQ) counterparts. LQ medical images often suffer from substantial diagnostic quality degradation due to suboptimal imaging conditions, such as reduced radiation exposure time and insufficient radiation intensity, which are employed to minimize potential health risks to patients. Significant progress has been made in recent years with specialized restoration models for specific MedIR tasks, including PET synthesis \cite{xiang2017xiang,chan2018dcnn,zhou2020cyclewgan,luo2022argan,zhou2022sgsgan,yang2023drmc}, CT denoising \cite{chen2017redcnn,liang2020edcnn,luthra2021eformer,wang2023ctformer,ozturk2024denomamba}, and MRI super-resolution \cite{yang2017dagan,huang2022swinmr,huang2022sdaut,sun2025funet}. 

Medical image restoration (MedIR) is a fundamental task in medical imaging, focused on reconstructing high-quality (HQ) images from their low-quality (LQ) counterparts. LQ medical images often suffer from substantial diagnostic quality degradation due to suboptimal imaging conditions, such as reduced radiation exposure time and insufficient radiation intensity, which are employed to minimize potential health risks to patients. Significant progress has been made in recent years for many specific MedIR tasks, including PET synthesis \cite{xiang2017xiang,chan2018dcnn,zhou2020cyclewgan,luo2022argan,yang2023drmc,zhou2022sgsgan,gong2024pet_diffusion}, CT denoising \cite{chen2017redcnn,liang2020edcnn,luthra2021eformer,wang2023ctformer,ozturk2024denomamba}, and MRI super-resolution \cite{yang2017dagan,huang2022swinmr,huang2022sdaut,sun2025funet}. 

Despite their success in specific scenarios, task-specific MedIR models face critical limitations that hinder their clinical adaptability. \textbf{(1) Limited Generalization}. In complex multimodal imaging workflows (e.g., PET/CT or PET/MRI), multiple MedIR tasks often coexist. However, due to inherent differences in imaging modalities and degradation types, task-specific models trained for one MedIR task struggle to adapt to others, leading to significant performance drops. \textbf{(2) Inefficiency and Redundancy}. Task-specific models require redundant development efforts and resource allocation, as each task demands separate architectures, training pipelines, storage solutions, and computational resources. This fragmented approach escalates costs and complicates clinical deployment. \textbf{(3) Data Scarcity and Isolation}. Task-specific models are particularly vulnerable to data scarcity, as they rely on narrow, task-specific datasets that are often limited in medical imaging. This constraint not only increases their susceptibility to model overfitting but also isolates them from potential cross-task and cross-modal synergies due to their task-specific training. In summary, task-specific models face limitations in generalization, efficiency, and data availability, which collectively hinder their scalability and real-world applicability.

To overcome the limitations of task-specific models, recent efforts \cite{li2022airnet,park2023discriminative_filter,ma2023prompted_CT,potlapalli2023promptir,yang2024amir} have focused on developing an All-in-One model capable of handling multiple tasks simultaneously. This approach directly addresses the three key limitations of task-specific models: (1) improving cross-task generalization through multitask training, (2) eliminating redundancies by consolidating workflows into a single model, and (3) mitigating data scarcity by leveraging both multitask and multimodal data. To handle various restoration tasks, recent All-in-One methods in natural image restoration use advanced techniques, such as contrastive learning \cite{li2022airnet} and visual prompting \cite{potlapalli2023promptir}, to learn task-discriminative representations that guide the model’s adaptation to different tasks. In medical image restoration, the pioneering work of AMIR \cite{yang2024amir} introduces a task routing strategy that allocates different tasks to separate network paths. These approaches have significantly contributed to the development of effective All-in-One models.

However, due to the substantial differences between MedIR tasks in terms of both modality and degradation types, using a shared model for such diverse tasks requires careful consideration of two crucial inter-task relationships: \textit{task interference}, which occurs when conflicting gradient update directions arise on the same parameter between tasks \cite{yang2024amir}, and \textit{task imbalance}, which refers to uneven optimization caused by varying learning difficulties inherent to each task \cite{kendall2018uncertainty}. Regarding \textit{task interference}, while several methods have attempted to address this \cite{park2023discriminative_filter,yang2024amir}, they still lack the adaptability to handle the complexity and diversity across different MedIR tasks. This is because most of them still rely on fixed parameters shared across tasks, which prevents them from adapting to the specific needs of each task. When gradient conflicts arise in these shared parameters, they inevitably lead to suboptimal performance. On the other hand, \textit{task imbalance} has been largely overlooked in existing studies. These methods fail to recognize that different MedIR tasks have varying levels of learning difficulty, and using a uniform weight for the loss across tasks often results in some tasks dominating while others remain undertrained. Therefore, there is a pressing need for novel All-in-One approaches that effectively address the inter-task relationships of \textit{task interference} and \textit{task imbalance} in MedIR tasks. 

In this paper, we introduce a novel task-adaptive transformer (TAT) that effectively addresses two crucial inter-task relationships—\textit{task interference} and \textit{task imbalance}—for All-in-One medical image restoration. This is accomplished through two key innovations: a task-adaptive weight generation strategy and a task-adaptive loss balancing strategy. Specifically, \textbf{(1)} to mitigate \textit{task interference} between distinct MedIR tasks, we propose a task-adaptive weight generation strategy that dynamically generates task-specific weight parameters for processing, thereby eliminating potential conflicts in weight updates. \textbf{(2)} To properly address \textit{task imbalance} and prevent task domination or undertraining, we introduce a task-adaptive loss balancing strategy that dynamically adjusts the loss weights for different tasks during training, ensuring the most effective optimization path. Extensive experiments demonstrate that our proposed TAT achieves state-of-the-art performance on tasks such as PET synthesis, CT denoising, MRI super-resolution, and All-in-One medical image restoration. 

% Our contributions are as follows:
% \begin{itemize} 
% \item We propose TAT, a novel task-adaptive transformer for All-in-One medical image restoration. TAT addresses the All-in-One challenge from a new perspective of managing inter-task relationships, including both \textit{task interference} and \textit{task imbalance}.
% \item We introduce a task-adaptive weight generation strategy to alleviate \textit{task interference} by generating task-specific weight parameters, thereby reducing conflicts during weight updates. 
% \item We propose a task-adaptive loss balancing strategy to address \textit{task imbalance} by dynamically adjusting the loss weights for different tasks based on their learning difficulty, thereby ensuring the most effective optimization route.
% \end{itemize} 

\section{Method}
In this section, we provide a detailed introduction to our proposed task-adaptive transformer (TAT). We first describe the overall architecture of TAT in Subsection~\ref{sec_overall}. Then, in Subsections~\ref{sec_weight_generation} and \ref{sec_task_balancing}, we present the two key innovations: the task-adaptive weight generation strategy, which addresses \textit{task interference}, and the task-adaptive loss balancing strategy, which addresses \textit{task imbalance}.

\subsection{Overall Network Architecture} \label{sec_overall}
\begin{figure*}[t]
	\centering
	\includegraphics[width=\textwidth]{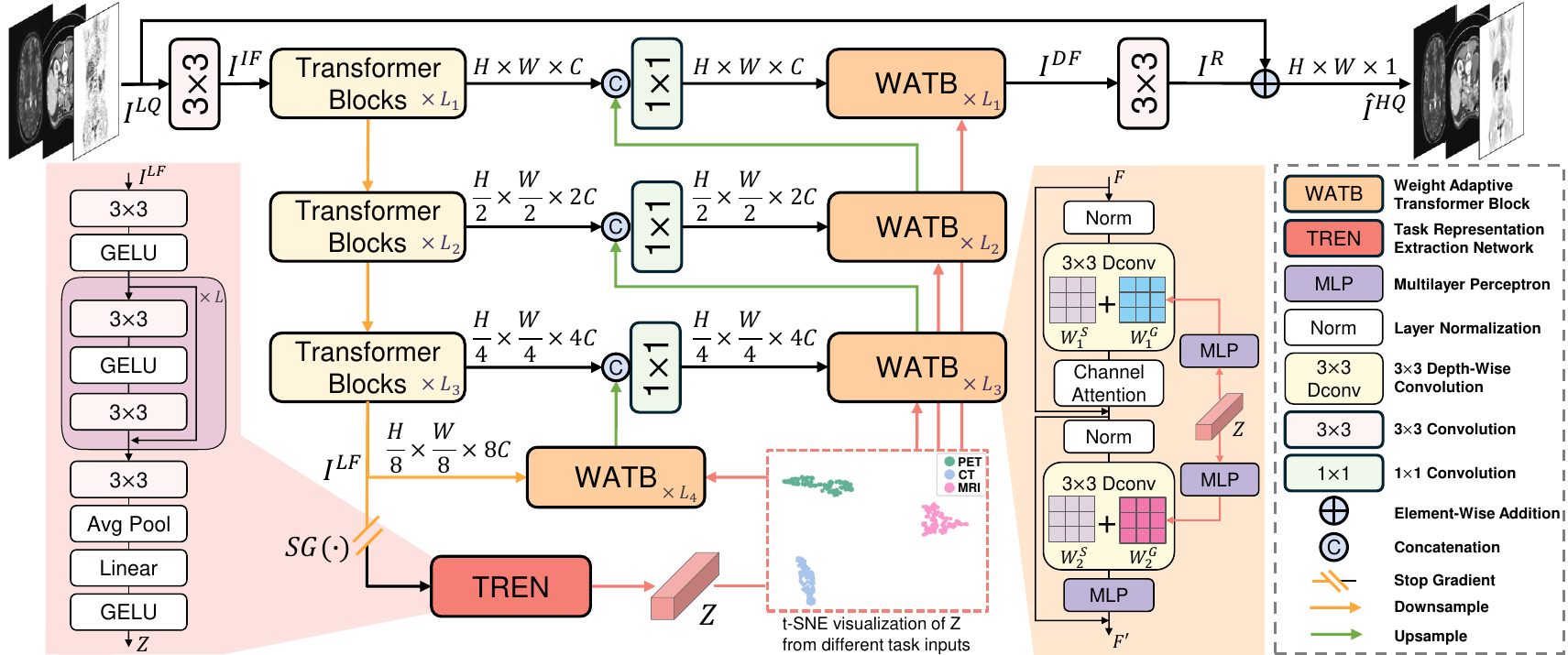}
    \caption{Overview of the proposed task-adaptive transformer (TAT) network.}
	\label{fig:framework}
\end{figure*}

As shown in Fig. \ref{fig:framework}, the proposed TAT features a multi-level U-shaped architecture consisting of a three-stage encoder with Transformer blocks \cite{zamir2022restormer} and a four-stage decoder incorporating weight-adaptive Transformer blocks (WATBs). To address the issue of \textit{task interference}, a task representation extraction network (TREN) is employed to extract task-specific representations that guide the generation of weights in the WATBs. The TAT begins by extracting initial features, denoted as $I^{IF} \in \mathbb{R}^{H \times W \times C}$, from an input low-quality (LQ) medical image $I^{LQ} \in \mathbb{R}^{H \times W \times 1}$ via a $3 \times 3$ convolutional layer, where $H$, $W$, and $C$ represent the height, width, and channel dimensions, respectively. These features are then encoded into latent representations $I^{LF}$ through the Transformer-based encoder. The pipeline subsequently splits into two branches: the first processes $I^{LF}$ through the decoder to produce deep features $I^{DF}$, while the second passes a gradient-detached copy of $I^{LF}$ through TREN to extract a task-specific representation $Z \in \mathbb{R}^{d}$. The WATBs utilize $Z$ to generate task-adaptive weights, enabling specialized feature refinement during the decoding process. Finally, a $3 \times 3$ convolutional layer transforms $I^{DF}$ into a residual image $I^{R} \in \mathbb{R}^{H \times W \times 1}$, which is added to the original LQ image $I^{LQ}$ to yield the restored high-quality (HQ) output, $\hat{I}^{HQ} = I^{LQ} + I^{R}$. 

\subsection{Task-Adaptive Weight Generation Strategy} 
\label{sec_weight_generation}
Most existing All-in-One models share a common limitation: they rely on a single model with fixed parameters to handle multiple tasks. This one-size-fits-all approach often results in suboptimal performance due to \textit{task interference}, where conflicting gradient updates from different tasks impede effective parameter optimization. As a result, the weight parameters are not specialized for any particular task, leading to diminished performance across broad. To address this, we propose a novel task-adaptive weight generation strategy that dynamically generates task-specific parameters for specialized processing, thereby eliminating potential interference. We introduce this strategy from two perspectives: task-specific representation extraction and task-adaptive weight generation.

\textbf{Task-Specific Representation Extraction.} Previous All-in-One models for natural images often utilize advanced techniques, such as contrastive learning \cite{li2022airnet} or auxiliary classification tasks \cite{park2023discriminative_filter}, to learn task-specific representations as guidance. However, we argue that these methods are unnecessary in the context of medical image restoration. Due to the significant semantic differences between various medical imaging modalities, the latent features encoded with semantic information inherently exhibit task-specific variations. Consequently, even straightforward feature extraction from the latent feature enables distinctions across tasks, eliminating the need for complex representation learning. Building upon this insight, we propose a simple task representation extraction network (TREN), consisting of sequential convolutional blocks that directly extract task-specific representations $Z\in \mathbb{R}^{d}$ from the latent features $I^{LF}$, as expressed below: 
\begin{equation}
Z=\operatorname{TREN}(\operatorname{SG}(I^{LF})),
\end{equation}
where $\operatorname{SG}(\cdot)$ denotes the stop-gradient operator, which decouples the extraction of latent features $I^{LF}$ from the extraction of task representations $Z$, thereby preventing potential interference between the two processes with distinct objectives. The resulting extracted task representations $Z$ are specific to each task, as evidenced by the t-SNE visualization in Fig.~\ref{fig:framework}.

\textbf{Task-Adaptive Weight Generation.} \textit{Task interference} occurs when different tasks conflict in their update directions for the same weight parameters. To address this, we propose generating task-specific parameters for each task. Using the task-specific representation $Z$, we use multi-layer perceptrons (MLPs) to estimate weight parameters for each decoding Transformer block. However, traditional approaches that generate weights for linear layers or standard convolutions face scalability issues: their parameter counts grow quadratically with channel dimension $C$ (\textit{i.e.}, $\mathcal{O}(C^2)$), leading to computational inefficiency and unreliable parameter estimation. To mitigate this, we shift focus to depth-wise convolutions, a lightweight alternative with only $k\times k\times C$ parameters (where 
$k$ is the kernel size), scaling linearly with $C$ (\textit{i.e.}, $\mathcal{O}(C)$) since $k \ll C$. This choice is motivated by two key advantages. First, depth-wise convolutions preserve local spatial information while complementing global attention mechanisms—a synergy shown to enhance performance in vision Transformers \cite{guo2022cmt,zamir2022restormer}. Second, their parameter efficiency enables accurate and compact weight generation. Consequently, our weight generation process can be formulated as follows:
\begin{equation}
W^G =\operatorname{Reshape}(\operatorname{MLP}(Z)),
\end{equation} 
where $W^G$ denotes the dynamically generated weight for depth-wise convolution, which is obtained by first transforming $Z$ through an MLP and then reshaping it into the target kernel shape. Finally, the generated task-specific weight $W^G$ is summed with the previously shared weight $W^S$ as follows:
\begin{equation}
W=W^S+\lambda W^{G},
\end{equation}
where $W$ is the final weight of the depth-wise convolution and $\lambda$ is a learnable parameter. By incorporating the generated task-specific weight into the Transformer block, we transform the original Transformer block \cite{zamir2022restormer} into a weight-adaptive Transformer block (WATB), as illustrated in Fig.~\ref{fig:framework}.

\subsection{Task-Adaptive Loss Balancing Strategy} 
\label{sec_task_balancing} 
Existing All-in-One methods ignore the challenge of \textit{task imbalance}, where different tasks present varying learning difficulties, leading to some tasks dominating while others remain undertrained. This concern has been addressed in the field of multi-task learning by using a loss balancing strategy \cite{kendall2018uncertainty} that dynamically allocates task-specific weights during training. A common formulation is:
\begin{equation}
Loss=\sum _{t=1}^{T}(\frac{1}{2\sigma _{t}^{2} }L^{t} + \log_{}{\sigma _{t}} ),
\label{eq_uncertainty}
\end{equation} 
where $T$ denotes the total number of tasks. $L^t$ denotes the loss for the $t$-th task. $\sigma_t \in \mathbb{R}^1$ is a learnable parameter. Here, $\frac{1}{2\sigma _{t}^{2} }$ dynamically scales the loss weight of each task, while $\log_{}{\sigma _{t}}$ regularizes the scaling. When the loss $L^t$ is large and tends to dominate the overall loss, $\sigma_t$ increases to suppress its weight, and vice versa. This mechanism autonomously balances task contributions, ensuring equitable training without manual intervention. However, while effective for task-level balancing, this approach lacks sample-level adaptability and struggles with implementation in task-specific models. To address these limitations, we propose a novel task-adaptive balancing strategy that achieves sample-level balancing by redefining the derivation of $\sigma \in \mathbb{R}^1$:
\begin{equation}
Loss=\frac{1}{2\sigma ^{2}}L_{1}(\hat{I}^{HQ},I^{HQ})+ \log_{}{\sigma} ,
\end{equation} 
\begin{equation}
\sigma = \operatorname{MLP}(\operatorname{SG}([L_1(I^{LQ},I^{HQ}),L_1(I^{LQ},\hat{I}^{HQ}),L_1(\hat{I}^{HQ},I^{HQ})])),
\end{equation} 
where $L_1(\cdot)$ denotes the L1 distance, and $\operatorname{SG}(\cdot)$ denotes the stop gradient operation that decouples loss balancing and model optimization. The three terms— $L_1(I^{LQ}, I^{HQ})$, $L_1(I^{LQ},\hat{I}^{HQ})$, and $L_1(\hat{I}^{HQ},I^{HQ})$—encodes sample-specific training dynamics. By feeding their concatenated values into an MLP, we estimate $\sigma$ adaptively for each sample, enabling fine-grained balancing. While this strategy shifts the derivation of $\sigma$ from task-index-conditioned ($\sigma_t$ in Eq.~\ref{eq_uncertainty}) to sample-loss-conditioned, the core mechanism—dynamic weighting via $\frac{1}{2\sigma^{2} }$ and regularization via $\log_{}{\sigma}$—remains aligned with the original theory, ensuring autonomous loss balancing while extending flexibility to the sample level.

\section{Experiments and Results} 
\subsection{Dataset}
We utilize the dataset provided by the paper \cite{yang2024amir}, which includes three distinct datasets from three different tasks: PET synthesis, CT denoising, and MRI super-resolution. (1) The PET synthesis dataset consists of paired low-dose LQ images, with a dose reduction factor of 12, and corresponding full-dose HQ images. Each image has a size of $400 \times 92$. The dataset includes 8,350 PET images for training, 684 for validation, and 2,044 for testing. (2) The CT denoising dataset consists of paired quarter-dose LQ images and corresponding standard-dose HQ images. Each image has a size of $512 \times 512$. The dataset includes 2,039 CT images for training, 128 for validation, and 211 for testing. (3) The MRI super-resolution dataset consists of paired $4 \times$ downsampled LQ images and corresponding HQ images. Each image has a size of $256 \times 256$. The dataset includes 40,500 MRI images for training, 5,828 for validation, and 11,400 for testing. 

\begin{table}[!ht] 
\centering
\caption{Task-specific medical image restoration results. The best results are \textbf{bolded}, and the second-best results are {\ul underlined}. * Denotes results that are significantly different from the best results paired t-test at $p<0.05$.} 
\resizebox{\textwidth}{!}{
\begin{tabular}{c|ccc|c|ccc|c|ccc}
\hline
\multirow{2}{*}{Method} & \multicolumn{3}{c|}{PET Synthesis}                 & \multirow{2}{*}{Method} & \multicolumn{3}{c|}{CT Denoising}                  & \multirow{2}{*}{Method} & \multicolumn{3}{c}{MRI Super-Resolution}            \\ \cline{2-4} \cline{6-8} \cline{10-12} 
                        & PSNR↑          & SSIM↑           & RMSE↓           &                         & PSNR↑          & SSIM↑           & RMSE↓           &                         & PSNR↑          & SSIM↑           & RMSE↓            \\ \hline
Xiang's \cite{xiang2017xiang}                 & 35.93$^{*}$    & 0.9167$^{*}$    & 0.0980$^{*}$    & REDCNN \cite{chen2017redcnn}                  & 33.19$^{*}$    & 0.9113$^{*}$    & 8.9427$^{*}$          & DAGAN \cite{yang2017dagan}                   & 30.55$^{*}$    & 0.9189$^{*}$    & 34.0866$^{*}$          \\ 
DCNN \cite{chan2018dcnn}                    & 36.27$^{*}$    & 0.9243$^{*}$    & 0.0954$^{*}$    & EDCNN \cite{liang2020edcnn}                   & 33.41$^{*}$    & 0.9155$^{*}$    & 8.7401$^{*}$    & SwinMR \cite{huang2022swinmr}                  & 30.93$^{*}$    & 0.9253$^{*}$    & 32.7339$^{*}$    
\\ 
CycleWGAN \cite{zhou2020cyclewgan}               & 36.62$^{*}$    & 0.9290$^{*}$    & 0.0910$^{*}$    & Eformer \cite{luthra2021eformer}                 & 33.35$^{*}$    & {\ul 0.9175$^{*}$}& 8.8030$^{*}$    & SDAUT \cite{huang2022sdaut}                   & 30.96$^{*}$    & 0.9257$^{*}$    & 32.5928$^{*}$    \\ 
ARGAN \cite{luo2022argan}                   & {\ul 36.73$^{*}$} & {\ul 0.9406$^{*}$} & {\ul 0.0902$^{*}$} & CTformer \cite{wang2023ctformer}                & 33.25$^{*}$    & 0.9134$^{*}$    & 8.8974$^{*}$    & F-UNet \cite{sun2025funet}                  & 31.26$^{*}$    & 0.9314$^{*}$    & 31.5675$^{*}$    \\ 
DRMC \cite{yang2023drmc}                    & 36.00$^{*}$    & 0.9352$^{*}$    & 0.0998$^{*}$    & DenoMamba \cite{ozturk2024denomamba}               & {\ul 33.53$^{*}$} & 0.9149$^{*}$& {\ul 8.6115$^{*}$} & MambaIR \cite{guo2025mambair}                 & {\ul 31.77$^{*}$} & {\ul 0.9369$^{*}$} & {\ul 29.8372$^{*}$} \\ 
TAT                     & \textbf{37.31\textcolor{white}{$^{*}$}} & \textbf{0.9482\textcolor{white}{$^{*}$}} & \textbf{0.0851\textcolor{white}{$^{*}$}} & TAT                     & \textbf{33.78\textcolor{white}{$^{*}$}} & \textbf{0.9199\textcolor{white}{$^{*}$}} & \textbf{8.3799\textcolor{white}{$^{*}$}} & TAT                     & \textbf{32.13\textcolor{white}{$^{*}$}} & \textbf{0.9408\textcolor{white}{$^{*}$}} & \textbf{28.8921\textcolor{white}{$^{*}$}} \\ \hline
\end{tabular}
}
\label{tab:single_task_comparison}
\end{table}  

%%%%%%%%%%%%%%%%%%%%%%%%%%%%%%%%%%%%%%%%%%%%%%%%%%%%%%%%%%%%%%%%%%%%%%%%%%
\begin{table}[!ht] 
\centering
\caption{All-in-One medical image restoration results. } 
\resizebox{\textwidth}{!}{
\begin{tabular}{c|ccc|ccc|ccc|ccc}
\hline
\multirow{2}{*}{Method}     & \multicolumn{3}{c|}{PET Synthesis}                 & \multicolumn{3}{c|}{CT Denoising}                  & \multicolumn{3}{c|}{MRI Super-Resolution}            & \multicolumn{3}{c}{Avg.}                        \\ \cline{2-4} \cline{5-8} \cline{9-13} 
           & PSNR↑          & SSIM↑           & RMSE↓           & PSNR↑          & SSIM↑           & RMSE↓           & PSNR↑          & SSIM↑           & RMSE↓            & PSNR↑          & SSIM↑           & RMSE↓           \\ \hline
ARGAN \cite{luo2022argan}      & 36.75$^{*}$    & 0.9389$^{*}$    & 0.0907$^{*}$    & 32.92$^{*}$    & 0.9111$^{*}$    & 9.2110$^{*}$    & 30.08$^{*}$    & 0.9083$^{*}$    & 35.7999$^{*}$    & 33.25$^{*}$    & 0.9194$^{*}$    & 15.0339$^{*}$   \\ 
DenoMamba \cite{ozturk2024denomamba}  & 36.81$^{*}$    & 0.9367$^{*}$    & 0.0895$^{*}$    & 33.18$^{*}$    & 0.9115$^{*}$    & 8.9512$^{*}$& 30.32$^{*}$    & 0.9091$^{*}$    & 34.6972$^{*}$    & 33.44$^{*}$    & 0.9191$^{*}$    & 14.5793$^{*}$   \\ 
MambaIR \cite{guo2025mambair}    & 37.17$^{*}$ & 0.9458$^{*}$    & 0.0864$^{*}$ & 33.50$^{*}$& 0.9165$^{*}$& 8.6345$^{*}$    & 31.31$^{*}$& 0.9305$^{*}$& 31.3150$^{*}$& 33.99$^{*}$    & 0.9309$^{*}$& 13.3453$^{*}$   \\ 
AirNet \cite{li2022airnet}     & 37.17$^{*}$ & 0.9451$^{*}$    & 0.0864$^{*}$ & 33.62$^{*}$& 0.9176$^{*}$    & 8.5226$^{*}$& 31.39$^{*}$    & 0.9316$^{*}$    & 31.1141$^{*}$    & 34.06$^{*}$    & 0.9314$^{*}$    & 13.2410$^{*}$   \\ 
AMIR \cite{yang2024amir}      & 37.12$^{*}$    & 0.9475$^{*}$ & 0.0876$^{*}$    & 33.70$^{*}$ & 0.9182$^{*}$ & 8.4520$^{*}$ & 32.03$^{*}$ & 0.9396$^{*}$ & 29.0988$^{*}$ & 34.28$^{*}$ & 0.9351$^{*}$ & 12.5461$^{*}$   \\ 
TAT        & \textbf{37.28}\textcolor{white}{$^{*}$} & \textbf{0.9480}\textcolor{white}{$^{*}$} & \textbf{0.0856}\textcolor{white}{$^{*}$} & \textbf{33.80}\textcolor{white}{$^{*}$} & \textbf{0.9192}\textcolor{white}{$^{*}$} & \textbf{8.3642}\textcolor{white}{$^{*}$} & \textbf{32.10}\textcolor{white}{$^{*}$} & \textbf{0.9402}\textcolor{white}{$^{*}$} & \textbf{28.9145}\textcolor{white}{$^{*}$} & \textbf{34.39}\textcolor{white}{$^{*}$} & \textbf{0.9358}\textcolor{white}{$^{*}$} & \textbf{12.4548}\textcolor{white}{$^{*}$} \\ \hline
\end{tabular}
}
\label{tab:all_in_one_comparison}
\end{table} 

\begin{table}[!ht] 
\centering
\caption{Ablation study results of TAT. } 
\resizebox{\textwidth}{!}{
\begin{tabular}{cc|c|ccc|ccc|ccc}
\hline
\multicolumn{2}{c|}{\multirow{2}{*}{Method}}                                  & \multirow{2}{*}{Params (M)} & \multicolumn{3}{c|}{PET Synthesis}                 & \multicolumn{3}{c|}{CT Denoising}                  & \multicolumn{3}{c}{MRI Super-Resolution}            \\ \cline{4-12} 
\multicolumn{2}{c|}{}                                                         &                             & PSNR↑          & SSIM↑           & RMSE↓           & PSNR↑          & SSIM↑           & RMSE↓           & PSNR↑          & SSIM↑           & RMSE↓            \\ \hline
\multicolumn{2}{c|}{TAT}                                                      & 41.69                       & \textbf{37.28}\textcolor{white}{$^{*}$} & \textbf{0.9480}\textcolor{white}{$^{*}$} & \textbf{0.0856}\textcolor{white}{$^{*}$} & \textbf{33.80}\textcolor{white}{$^{*}$} & \textbf{0.9192}\textcolor{white}{$^{*}$} & \textbf{8.3642}\textcolor{white}{$^{*}$} & \textbf{32.10}\textcolor{white}{$^{*}$} & \textbf{0.9402}\textcolor{white}{$^{*}$} & \textbf{28.9145}\textcolor{white}{$^{*}$} \\ \hline
\multicolumn{1}{c|}{\multirow{3}{*}{Weight Generation}} & w/o                 & 26.12                       & 37.16$^{*}$          & 0.9472$^{*}$          & 0.0871$^{*}$          & 33.64$^{*}$          & 0.9181$^{*}$          & 8.4845$^{*}$          & 31.85$^{*}$          & 0.9374$^{*}$          & 29.7689$^{*}$          \\
\multicolumn{1}{c|}{}                                   & Generate All Params & 663.14                      & 37.20$^{*}$          & 0.9455$^{*}$          & 0.0868$^{*}$          & 33.69$^{*}$          & 0.9183$^{*}$          & 8.4611$^{*}$          & 31.89$^{*}$          & 0.9382$^{*}$          & 29.5431$^{*}$          \\
\multicolumn{1}{c|}{}                                   & w/o Stop Gradient   & 41.69                       & 37.25\textcolor{white}{$^{*}$}          & 0.9469$^{*}$          & 0.0860\textcolor{white}{$^{*}$}          & 33.69$^{*}$          & 0.9188$^{*}$          & 8.4559$^{*}$          & 32.07$^{*}$          & 0.9401\textcolor{white}{$^{*}$}          & 29.0291$^{*}$          \\ \hline
\multicolumn{1}{c|}{\multirow{2}{*}{Loss Balancing}}    & w/o                 & 41.69                       & 37.11$^{*}$          & 0.9474$^{*}$          & 0.0877$^{*}$          & 33.72$^{*}$          & 0.9183$^{*}$          & 8.4338$^{*}$          & 32.01$^{*}$          & 0.9397$^{*}$          & 29.1954$^{*}$          \\
\multicolumn{1}{c|}{}                                   & Task-Level \cite{kendall2018uncertainty}          & 41.69                       & 37.22$^{*}$          & 0.9475$^{*}$          & 0.0865$^{*}$          & 33.76$^{*}$          & 0.9191\textcolor{white}{$^{*}$}          & 8.3845$^{*}$          & 32.05$^{*}$          & 0.9398$^{*}$          & 29.0858$^{*}$          \\ \hline
\end{tabular}
}
\label{tab:ablation}
\end{table}

\begin{figure*}[t]
	\centering
	\includegraphics[width=\textwidth]{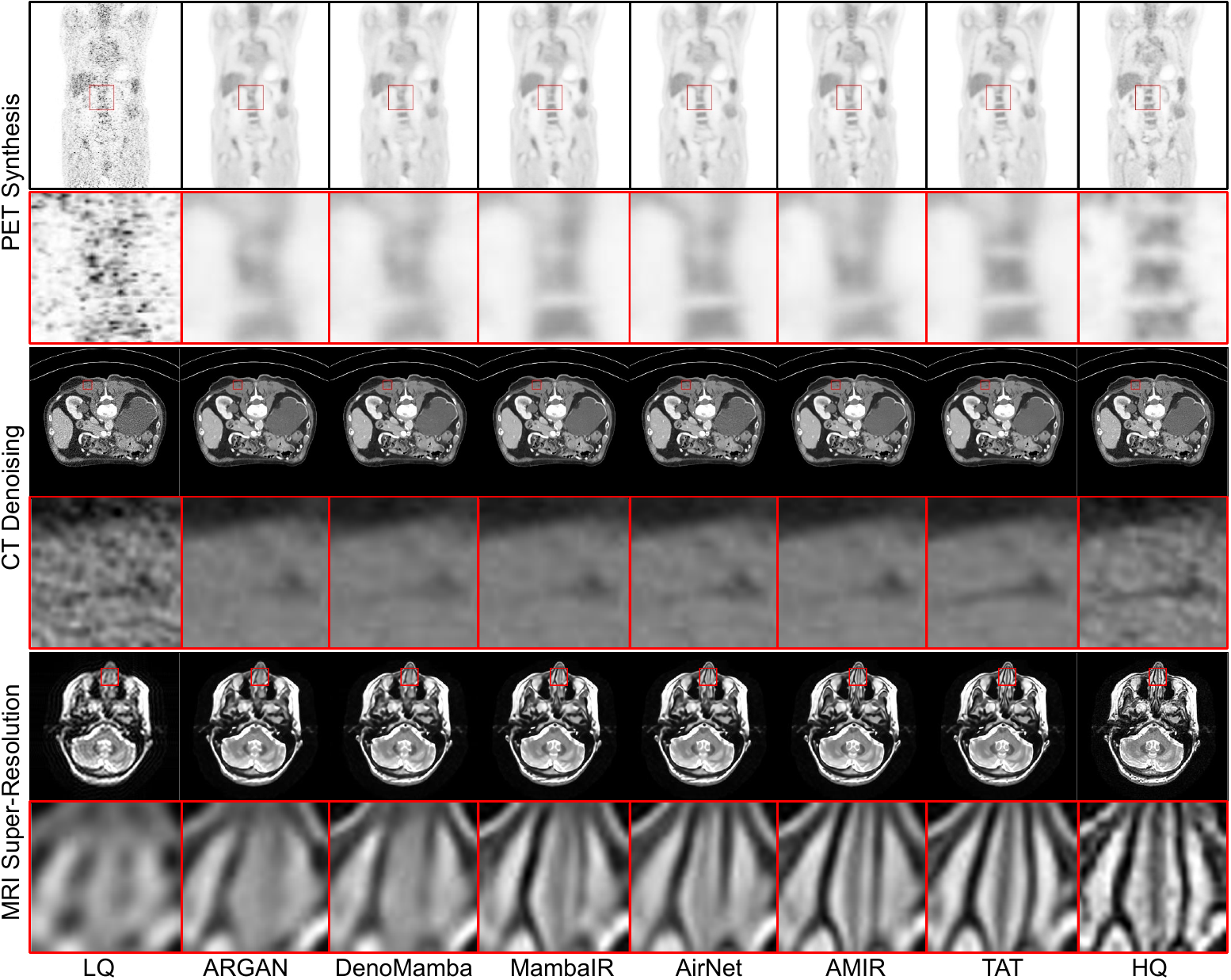}
    \caption{Visual comparisons on All-in-One medical image restoration.}
	\label{fig:visual comparison}
\end{figure*} 

\subsection{Implementation}
For model architecture, the numbers of feature extraction blocks in TAT are $L_1=4$, $L_2=L_3=6$, and $L_4=8$. The residual block number in TREN is $L=2$. The task-specific representation $Z$ has a dimensionality of $d=256$. For model training, we use a total batch size of 12 (4 samples per dataset) and a patch size of $128 \times 128$. The model is optimized using the AdamW optimizer with a learning rate of $2 \times 10^{-4}$ trained for $4 \times 10^{5}$ iterations. For model evaluation, the restoration performance is evaluated using PSNR, SSIM, and RMSE metrics.

\subsection{ Comparative Experiment}
For task-specific MedIR, we compare TAT with five state-of-the-art (SOTA) methods for each individual task. For All-in-One MedIR, we compare TAT with five methods: two SOTA methods (AirNet \cite{li2022airnet} and AMIR \cite{yang2024amir}) designed for All-in-One restoration, and three methods (ARGAN \cite{luo2022argan}, DenoMamba \cite{ozturk2024denomamba}, and MambaIR \cite{guo2025mambair}) that perform the best in their respective specific tasks.

\textbf{Task-Specific Medical Image Restoration.}
Although TAT is not specifically designed for task-specific MedIR, it still achieves the best performance as shown in Table.~\ref{tab:single_task_comparison}. This suggests that the two proposed task-adaptive strategies are extensible and effective for individual tasks as well.

\textbf{All-in-One Medical Image Restoration.}
The results of All-in-One MedIR are presented in Table.~\ref{tab:all_in_one_comparison}, where TAT significantly ($p<0.05$) outperforms all comparison methods across all three tasks. Especially, it outperforms AMIR which is the current SOTA method in All-in-One MedIR. This is because TAT best deals with the inter-task relationships of \textit{task interference} and \textit{task imbalance}. In fact, our proposed TAT even achieves a comparable performance to task-specific models present in Table.~\ref{tab:single_task_comparison}. This indicates that TAT has strong model capacity and adaptability to deal with diverse tasks. The visual comparisons in Fig.~\ref{fig:visual comparison} further highlight TAT's superior ability to consistently restore finer structures and details of images across different tasks.

\subsection{Ablation Study}
We conduct comprehensive ablation studies by systematically removing or modifying individual components. The experimental results are summarized in Table.~\ref{tab:ablation}. For the task-adaptive weight generation, we introduce model variants: (1) without the weight generation strategy, (2) with the generation of all parameters in the Transformer block, and (3) without the stop-gradient operation. All variants exhibit consistent performance degradation, highlighting the validity of our design choices. For the task-adaptive loss balancing strategy, we evaluate two model variants: (1) without loss balancing and (2) with a conventional task-level balancing strategy \cite{kendall2018uncertainty}. Both alternatives lead to significant performance degradation, further validating the superiority of our loss balancing strategy.

\section{Conclusion}
In this paper, we introduce a novel task-adaptive transformer (TAT) to address the challenges of \textit{task interference} and \textit{task imbalance} in All-in-One medical image restoration. To mitigate \textit{task interference}, we propose a task-adaptive weight generation strategy that produces task-specific weight parameters, thereby reducing conflicts during weight updates. To tackle \textit{task imbalance}, we introduce a task-adaptive loss balancing strategy that dynamically adjusts the loss weights according to the learning difficulty of each task, ensuring the most effective optimization path. Experimental results demonstrate that our approach achieves state-of-the-art performance across different tasks. Considering that the two proposed task-adaptive strategies of TAT are architecture-agnostic, we plan to explore applying them to more modern architectures \cite{yang2024rat,yang2024restore_rwkv,guo2025mambair} in future work.

\begin{credits}
\subsubsection{\ackname} 
This work is supported by the Beijing Natural Science Foundation under Grant QY24144, the National Natural Science Foundation in China under Grant 62371016 and U23B2063, the Bejing Natural Science Foundation Haidian District Joint Fund in China under Grant L222032, the Fundamental Research Funds for the Central University of China from the State Key Laboratory of Software Development Environment in Beihang University in China, and the 111 Proiect in China under Grant B13003, and the high performance computing resources at Beihang University.

\subsubsection{\discintname}
We have no conflicts of interest to disclose.
\end{credits}

%%%%%%%%%%%%%%%%%%%%%%%%%%%%%%%%%%%%%%%%%%%%%%%%%%%%%%%%%%%%%%%%%%%%%%%%%%%%%%%%%

\bibliographystyle{splncs04}
\bibliography{Paper-1155}

\end{document}